\def\year{2019}\relax
\documentclass[sigconf]{acmart}
\usepackage{times}  
\usepackage{helvet}  
\usepackage{courier}  
\usepackage{url}  
\usepackage{graphicx}  
\usepackage[utf8]{inputenc} 
\usepackage[T1]{fontenc}    
\usepackage{hyperref}       
\usepackage{url}            
\usepackage{booktabs}       
\usepackage{amsfonts}       
\usepackage{nicefrac}       
\usepackage{microtype}      
\usepackage{subcaption}      
\usepackage{multirow,array}
\usepackage{float}

\usepackage{graphicx} 
\usepackage{amsmath}
\usepackage{amssymb}

\newtheorem{defn}{Definition}
\newtheorem{thm}{Theorem}

\newtheorem{claim}{Result}

\copyrightyear{2019} 
\acmYear{2019} 
\setcopyright{acmcopyright}
\acmConference[AIES '19]{2019 AAAI/ACM Conference on AI, Ethics, and Society}{January 27--28, 2019}{Honolulu, HI, USA}
\acmBooktitle{2019 AAAI/ACM Conference on AI, Ethics, and Society (AIES'19), January 27--28, 2019, Honolulu, HI, USA}
\acmPrice{15.00}
\acmDOI{10.1145/3306618.3314268}
\acmISBN{978-1-4503-6324-2/19/01} 
\title{Learning Existing Social Conventions via Observationally Augmented Self-Play}

\author{Adam Lerer}
\authornote{Both authors contributed equally to this research.}
\affiliation{%
  \institution{Facebook AI Research}
}
\email{alerer@fb.com}

\author{Alexander Peysakhovich}
\authornotemark[1]
\affiliation{%
  \institution{Facebook AI Research}
}
\email{alexpeys@fb.com}

\renewcommand{\shortauthors}{Lerer and Peysakhovich}

\begin{document}

\begin{abstract}
In order for artificial agents to coordinate effectively with people, they must act consistently with existing conventions (e.g. how to navigate in traffic, which language to speak, or how to coordinate with teammates). A group's conventions can be viewed as a choice of equilibrium in a coordination game. We consider the problem of an agent learning a policy for a coordination game in a simulated environment and then using this policy when it enters an existing group. When there are multiple possible conventions we show that learning a policy via multi-agent reinforcement learning (MARL) is likely to find policies which achieve high payoffs at training time but fail to coordinate with the real group into which the agent enters. We assume access to a small number of samples of behavior from the true convention and show that we can augment the MARL objective to help it find policies consistent with the real group's convention. In three environments from the literature - traffic, communication, and team coordination - we observe that augmenting MARL with a small amount of imitation learning greatly increases the probability that the strategy found by MARL fits well with the existing social convention. We show that this works even in an environment where standard training methods very rarely find the true convention of the agent's partners.
\end{abstract}

\maketitle

\section{Introduction}
A ubiquitous feature of social interaction is a need for individuals to coordinate \cite{lewis2008convention,shoham1997emergence,bicchieri2005grammar,stone2010ad,barrett2017making}. A common solution to the coordination problem is the establishment of social conventions which control daily tasks such as choosing which side of the road to drive on, who should get right of way during walking, what counts as polite, what language to speak, or how a team should apportion tasks. If we seek to construct artificial agents that can coordinate with humans, they must be able to act according to existing conventions.

In game theory, Nash equilibria are strategies for all players such that if everyone behaves according to them no individual can improve their payoff by deviating. In game theoretic models a convention is one of multiple possible equilibria in a coordination game \cite{lewis2008convention}. Stated in these terms our agent's task is to construct a policy that does well when paired with the equilibrium being played by existing agents.

There has been great recent progress in constructing policies that can do well in both single and multi-agent environments using deep reinforcement learning \cite{mnih2015human,silver2017mastering}. Deep RL methods typically require orders of magnitude more experience in an environment to learn good policies than humans \cite{lake2017building}, so agents are typically trained in simulation before being deployed onto the real task.

In zero-sum two-player environments (e.g. Go), it is the policy of the other player that is simulated during training. Typically, policies for both players are trained simultaneously or iteratively, in a process called self-play. Self-play produces successful policies because if self-play converges then it converges to an equilibrium of the game \cite{fudenberg1998theory} and in two-player, zero-sum games all equilibria are minimax/maximin strategies \cite{neumann1928theorie}. Thus, a fully converged strategy is guaranteed to be unexploitable for the task of interest (e.g. playing Go with a human champion). 

Constructing agents that can cooperate and coordinate with each other to achieve goals (e.g. work together as team to finish a task) has been a long running topic in multi-agent reinforcement learning (MARL) research \cite{walker1995understanding,stone2001scaling,kapetanakis2002reinforcement,lowe2017multi}. However, this literature typically assumes that the cooperating agents will continue to interact with those with whom they have been co-trained (this is sometimes referred to as ``centralized training with distributed execution''). In this case, if MARL converges, it finds an equilibrium and since agents will play with the same partners they trained with they will achieve these equilibrium payoffs. 

Unfortunately, agents are no longer guaranteed equilibrium payoffs if there are multiple equilibria and agents must coordinate with those they were \textit{not} trained with (in other words, when we remove the centralized training assumption). For example, training self-driving cars in a virtual environment may lead to agents that avoid crashing into each other during training, but drive on the wrong side of the road relative to the society they will enter.

In this paper we propose to give the agent access to a small amount of observations of existing social behavior, i.e. samples of (state, action) pairs from the test time environment. We focus on how such data, though it is not enough to purely clone good policies, can be used in the training process to learn the best response to the policies of the future partners. Our key assumption is that the existing environment already has some existing, stable, social conventions. Thus, our assumption is that future partners will be playing \textit{some} equilibrium strategies. In simple environments, agents could simply enumerate all possible equilibria and choose the one most consistent with the data. However, in more complex environments this becomes intractable. We propose to guide self-play toward the correct equilibrium by training with a joint MARL and behavioral cloning objective. We call this method observationally augmented self-play (OSP).

We consider OSP in several multi-agent situations with multiple conventions: a multi-agent traffic game \cite{sukhbaatar2016learning,resnick2018vehicle}, a particle environment combining navigation and communication \cite{mordatch2017emergence,lowe2017multi} and a Stag Hunt game where agents must take risks to accomplish a joint goal \cite{yoshida2008game,peysakhovich2017prosocial}. In each of these games we find that self-play can converge to multiple, incompatible, conventions. We find that OSP is able to learn correct conventions in these games with a small amount of observational data. Our results in the Markov Stag Hunt show that OSP can learn conventions it observes even when those conventions are very unlikely to be learned using MARL alone.

We do not claim that OSP is the ultimate approach to constructing agents that can learn social conventions. The success of OSP depends on both the game and the type of objective employed, thus an exploration of alternative algorithms is an important direction for future work. Our key result is that the combination of (a) a small number of samples from trajectories of a multi-agent game, and (b) knowledge that test time agents are playing some equilibrium gives much stronger test time performance than either component alone.

\section{Related Work}
OSP is related to existing work on adding reward shaping in MARL \cite{kapetanakis2002reinforcement,kapetanakis2002improving,babes2008social,devlin2011theoretical}. However the domain of interest differs slightly as reward shaping is typically used to cause all agents in a group to converge to a high-payoff equilibrium whereas we are interested in using shaping to guide training to select the correct test-time equilibrium.

Recent work has pointed out that strategies learned via a single instance of independent MARL can overfit to other agents' policies during training \cite{lanctot2017unified}. This work differs from ours in that it suggests the training of a single best response to a mixture of heterogeneous policies. This increases the robustness of agent policies but does not solve the problem of multiple, incompatible conventions that we study here. 

The approach of combining supervised learning from trajectories with RL has been studied in the single agent case \cite{hester2017learning}. In that work the trajectories are expert demonstrations and are used to guide RL to an optimal policy. In this case supervision is used to speed up learning while in our work the trajectories are used to select among many possible optima (equilibria) which may be equivalent at training time but not at test time. However this literature explores many methods for combining imitation learning and RL and some of these techniques may be interesting to consider in the multi-agent setting.

\section{Conventions in Markov Games}
A partially observed Markov game $G$ \cite{shapley1953stochastic,littman1994markov} consists of a set of players $P = \lbrace 1, \dots, N \rbrace$, a set of states $\mathcal{S}$, a set of actions for every player $A_i$ with the global set $\mathcal{A} = \times_{i \in P} A_i$, a transition function $\mathcal{\tau}: \mathcal{S} \times \mathcal{A} \to \Delta \mathcal{S}$, a reward function for each player that takes as input the current state and actions $R_i: \mathcal{S} \times \mathcal{A} \to \mathbb{R}.$ Players have observation functions $O_i: \mathcal{S} \to \mathcal{O}_i$ and can condition their behavior on these observations. Markov policies for each player are functions $\pi_i: \mathcal{O}_i \to A_i.$ Let $\Pi_i$ denote the set of all policies available to a player and $\Pi = \times_{i \in P} \Pi_i$ be the set of joint policies.\footnote{We consider only Markov policies in this work so that we can work with individual state-action pairs, although our same approach could be applied across observed trajectories to learn non-Markov policies (i.e. policies conditioned on their full history).}

We use the standard notation $\pi_{-i}$ to denote the policy vector for all players other than $i$. A set of policies $\pi \in \Pi$ and a (possible random) initial state $s_0 \in \mathcal{S}$ defines a (random) trajectory of rewards for each player. We let the value function $V_i (s, \pi_i, \pi_{-i})$ denote the discounted expectation of this trajectory. The best response starting at state $s$ for player $i$ to $\pi_{-i}$ is $BR(s, \pi_{-i}) = \text{argmax}_{\pi_i \in \Pi_i} V_i(s, \pi_i, \pi_{-i}).$ We let $s_0$ be the (possibly random) initial state of the game. 

There are many ways to extend the notion of a Nash equilibrium to the case of stochastic games. We will consider the Markov perfect best response. We denote by $BR(\pi_{-i})$ the policy (or policies) which is a best response starting at \textit{any} state and consider equilibria to be policies for each player $\pi$ such that each $\pi_i \in BR(\pi_{-i})$.\footnote{There are weaker notions, for example, requiring that policies are best responses at every state reached during play. It is known that Markov perfect equilibria are harder to find by learning \cite{fudenberg1998theory} and it is interesting to consider whether different kinds of choices (e.g. on-policy vs. off-policy learning) can make stronger or weaker convergence guarantees. However, these questions are outside the scope of this paper.}

We consider games with multiple, incompatible, conventions. Formally we say that conventions (equilibrium policy sets) $\pi$ and $\pi'$ are incompatible if the compound policy $(\pi_{i}, \pi'_{-i})$ is not an equilibrium.

\begin{figure}[h!]
\begin{center}
\includegraphics[scale=.33]{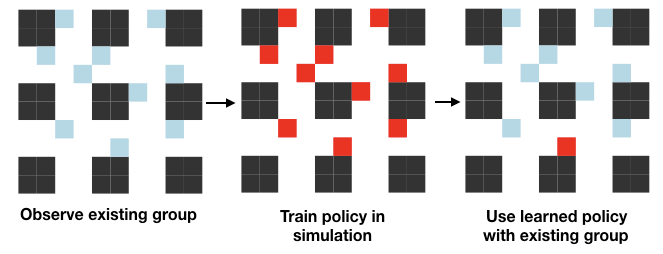}
\end{center}
\caption{A graphical description of our problem. Our central question is how to use observational data to make sure that training converges to the same convention as the group which our agent will enter.} 
\label{problem_explain}
\end{figure}

The goal of training is to compute a policy $\pi_i$ which is a best response to the existing convention $\pi^{test}_{-i}$. During training, the agent has access to the environment but receives only a limited set of observations of $\pi^{test}$ in the form of a set $\mathcal{D}$ of state-action pairs sampled from $\pi^{test}$. This is summarized in Figure \ref{problem_explain}.

We denote a generic element of $\mathcal{D}$ by $d = (s, a_k)$ which is a (state, action) pair for agent $k$.  Let $\mathcal{D}_j$ denote the subset of $\mathcal{D}$ which includes actions for agent $j$.

The dataset $\mathcal{D}$ may be insufficient to identify a best response to all possible policies $\pi^{test}$ consistent with $\mathcal{D}$. However, the set of equilibrium policy sets is typically much smaller than all possible policy sets. Therefore, if we assume that all agents are minimizing regret then we must only consider equilibrium policy sets consistent with $\mathcal{D}$.

Given a game $G$ and dataset $\mathcal{D}$, a brute force approach to learn a policy compatible with the conventions of the group the agent will enter would be to compute the equilibrium $\pi^*$ of $G$ that maximizes the likelihood of $\mathcal{D}$. Formally this is given by 
\begin{align*}
	\pi^* \in \text{argmax}_{\pi} \sum_i \sum_{(s,a)\in \mathcal{D}_i} \log(\pi_i(s, a)) \\  \text{ s.t. } \forall i,\ \pi_i \in BR(\pi_{-i}).
\end{align*}
This constrained optimization problem quickly becomes intractable and we will instead try to find an equilibrium using multi-agent learning, and use $\mathcal{D}$ to increase the probability learning converges to it.

\subsection{Observationally Initialized Best Response Dynamics}
To get an intuition about how data can be used during training we will first study a learning rule where analytic results are more easily obtained. We will begin with the simplest multi-agent learning algorithm: \textit{best response dynamics} in a $2$ player Markov game where equilibria are in pure policies and are incompatible. 

In best response dynamics each player begins with a policy initialized at $\pi_i^0$. Players alternate updating their policy with the best response $\pi_i^{t} = BR(\pi_{-i}^{t-1}).$ When there are multiple best responses we assume there is a (non-randomized) tie-breaking rule used to select one. Given an equilibrium $A$ we denote by $\Pi^A$ the basin of attraction of $A$ (the set of initial states from which BR dynamics converge to $A.$) 

A naive way to use the observed data is to force the policies be consistent with $\mathcal{D}$ at each step of the best response dynamic by changing it at each states where it differs. However, this can introduce new equilibria to the game.\footnote{For a simple example consider a game where $5$ agents choose an action $A$ or $B$ and receive reward equal to the number of other agents whose actions match theirs. In this case there are equilibria where all agents choose $A$ or all agents choose $B$. If we restrict one agent to always choose $A$ we can introduce a new equilibrium where $4$ agents choose $B$ and one agent chooses $A$.}  

In the context of reward shaping it is well known that the way to avoid the introduction of new equilibria is to use potential based reward shaping \cite{devlin2011theoretical}, or, equivalently, use our information to only change the initialization of learning \cite{wiewiora2003potential}. We will follow this advice and study \textit{observationally initialized best response dynamics}. We begin with a policy $\pi_0$ chosen at random. However, for every player $i$ and state-action pair $(s, a)$ in the data $\mathcal{D}$ we form $\bar{\pi}^0$ by setting the corresponding action of $\pi^0_i (s)$ to $a$. We then perform best response dynamics from this new initialization. 

We will now discuss a class of games for which we can guarantee that observationally initialized best response dynamics have a larger basin of attraction for the equilibrium from which $\mathcal{D}$ was drawn relative to standard best response dynamics. This class is a generalization of a commonly used matrix game class: games with \textit{strategic complements} \cite{bulow1985multimarket}. For our purposes strategic complements corresponds to assuming that one's partners behave more consistently with some convention then one is also more incentivized to behave according to that convention.\footnote{In economic applications the notion of strategic complements is utilized in production games and roughly corresponds either the the idea of network effects (the more people use some product the higher a consumer's utility is from using that product) or a joint monotonicity condition (if Firm X produces cars and firm Y produces car washing materials if firm X produces more cars then firm Y sees higher demand for car washing materials). See the Supplement for a more formal discussion.} In existing work strategic complements are defined with respect to a single action rather than a Markov policy. To generalize to Markov games we introduce a notion of distance between policies:

\begin{defn}[Policy Closeness]
Given a player $i$ and target policy $A$ we say that policy $\pi$ is \textbf{weakly closer to $A$ than policy $\pi'$} if on all states either $\pi(s) = \pi'(s)$ or $\pi(s) = A(s)$. We denote this by $\pi \succsim_{A} \pi'.$
\end{defn}

Policy closeness gives us a partial ordering on policies which we use to generalize the idea of strategic complements.

\begin{defn}[Markov Strategic Complements]
A Markov game exhibits \textbf{Markov strategic complements} if for any equilibrium $A = (A_1, A_2)$ we have that $\pi_{i} \succsim_{A_i} \pi_i'$ implies that $BR(\pi_i) \succsim_{A_{-i}} BR(\pi_{i}').$ \end{defn}

Let $\mathcal{D}_A$ be a dataset drawn from equilibrium $A$ by sampling states and their equilibrium actions. Let $\bar{\Pi}^A (\mathcal{D}_A)$ be the basin of attraction of $A$ given observationally initialized best response dynamics. 

\begin{thm}
If a game where best-response dynamics always converge exhibits Markov strategic complements then for any $\mathcal{D}_A$  drawn from a equilibrium $A$ $\Pi^A \subseteq \bar{\Pi}^A (\mathcal{D}_A)$ and there exists $(s,A_i(s))$ such that if $(s,A_i(a)) \in \mathcal{D}_A$ then $\Pi^A \subset \bar{\Pi}^A (\mathcal{D}_A).$
\end{thm}

We relegate the proof to the Appendix. Roughly, it has two steps: first, we show that if $\pi_0$ is in the basin of attraction of $A$ then anything closer to $A$ is also in the basin. Second, we show that there is an initial state that is not in the basin of attraction of best response dynamics but is in the basin of attraction of observationally initialized best response dynamics. Because initialization can increase the basin of attraction without introducing any new equilibria the observed data can strictly improve the probability that we learn a policy consistent with the observed agents.

\section{Experiments}

\subsection{Observationally Augmented Self-Play}

\begin{figure*}[ht!]
\begin{center}
\includegraphics[scale=.5]{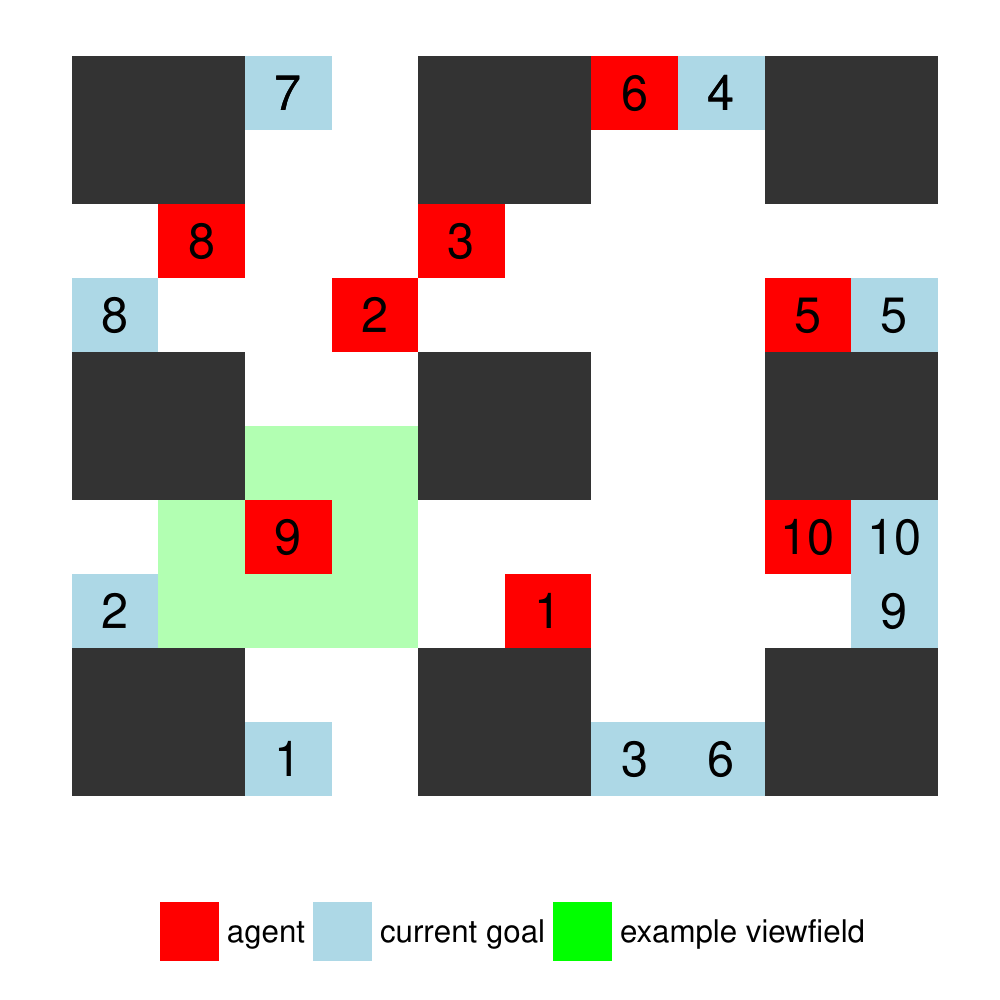}
\includegraphics[scale=.4]{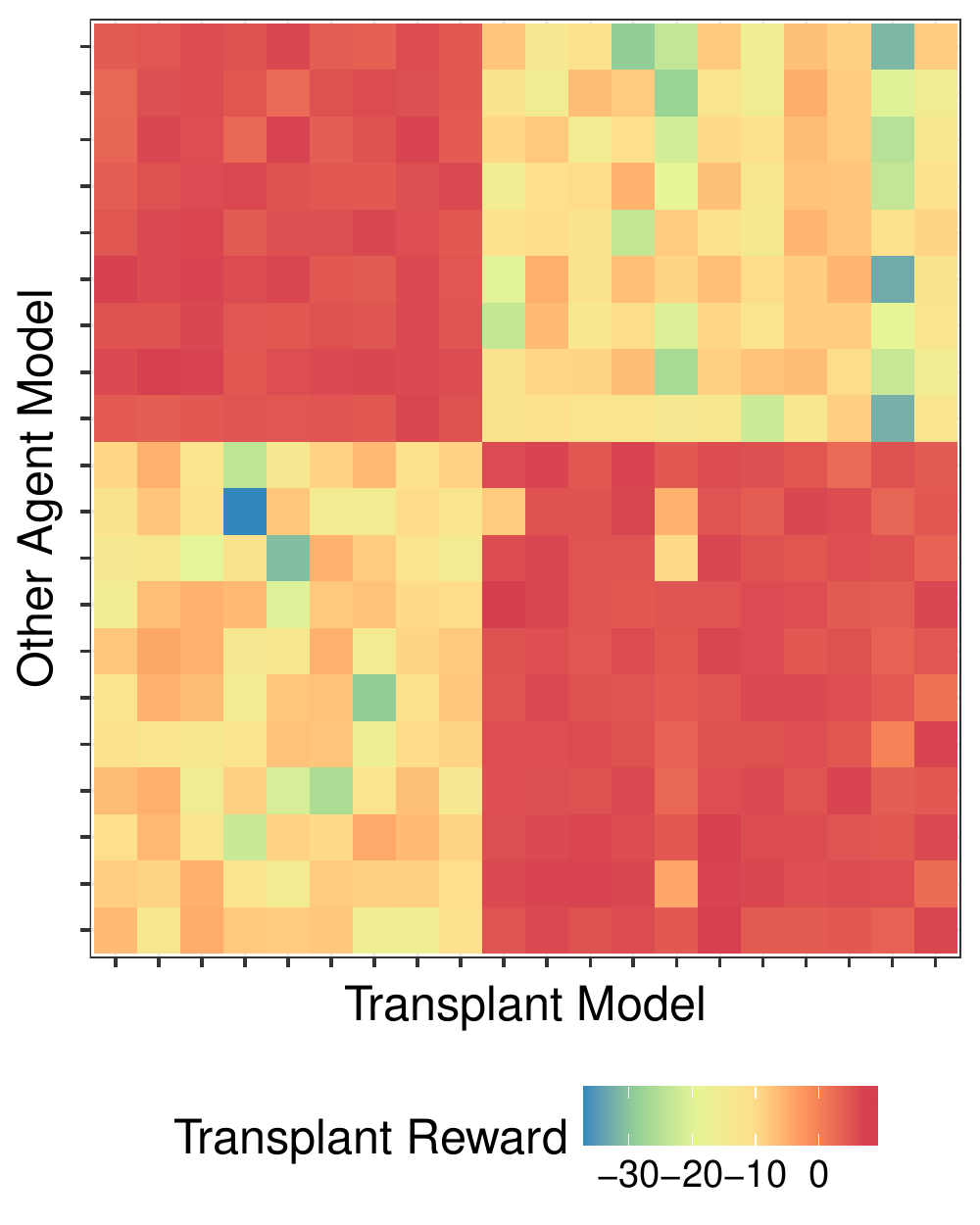} \\
\includegraphics[scale=.4]{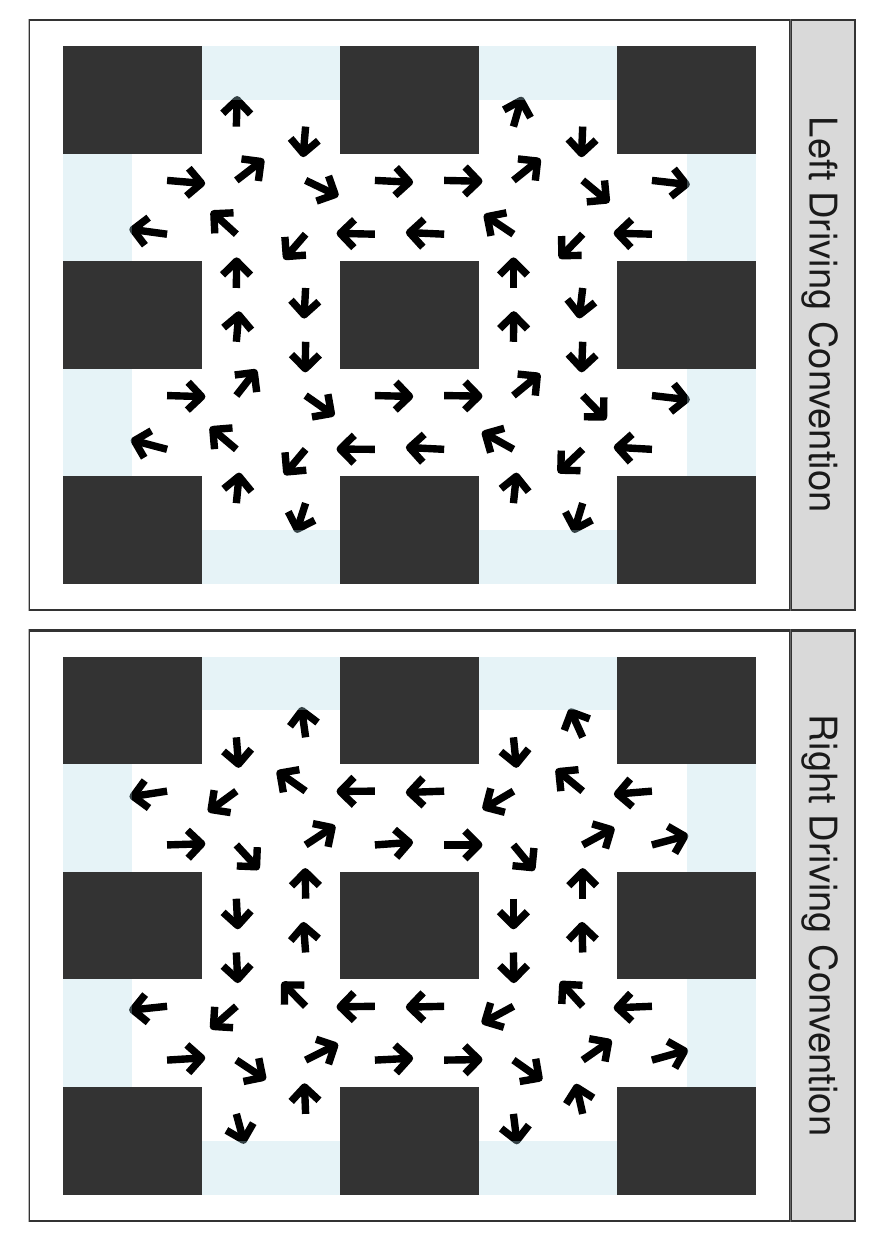}
\includegraphics[scale=.4]{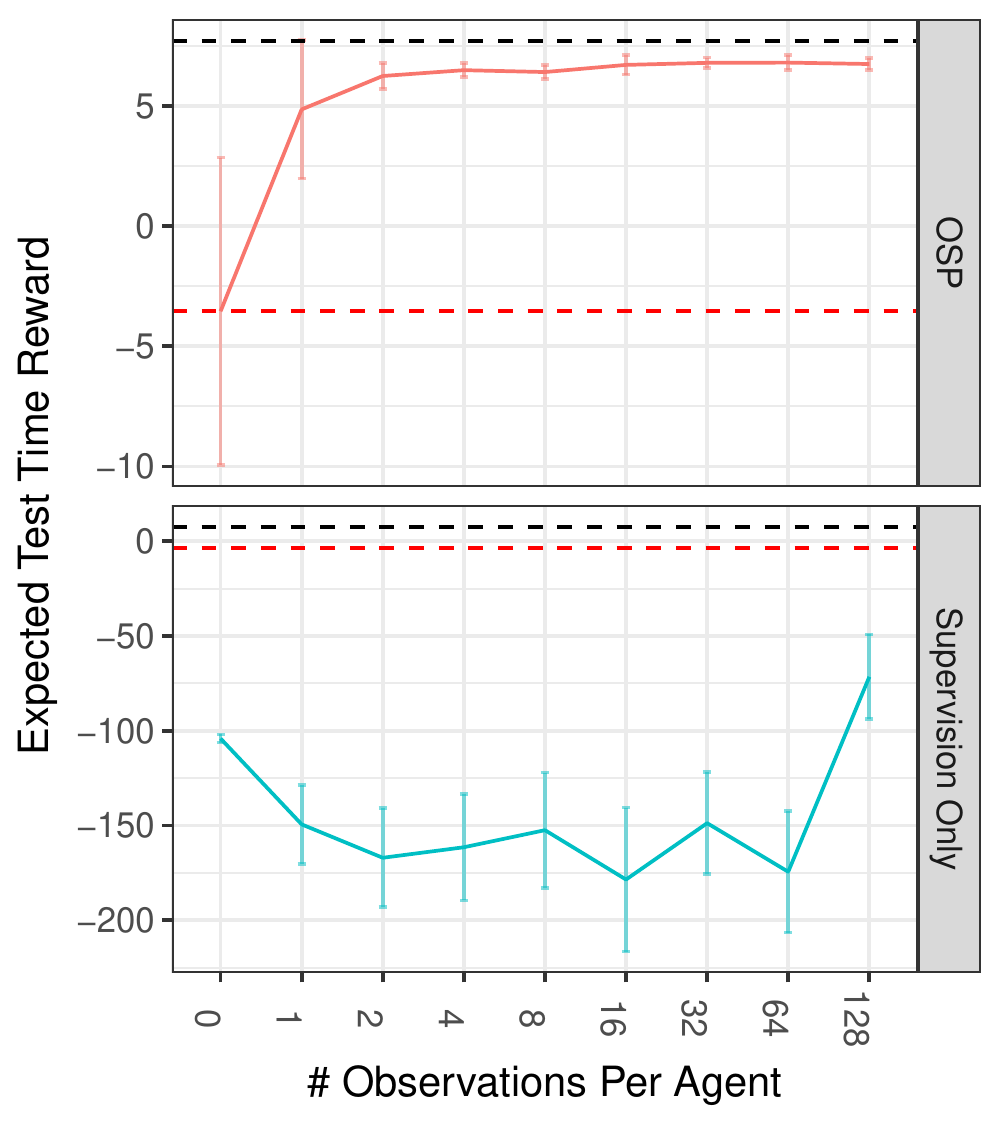}
\caption{In the traffic game 10 agents navigate a grid world to reach goals (first panel). Agents have a local field of view (indicated in green). Two incompatible conventions emerge across multiple training runs. MARL agents achieve high rewards with their training partners (diagonal) but lower rewards with incompatible partners (second panel). Visualizing these conventions shows that they differ in which side of the road to drive on (third panel). Given few observations for each agent OSP learns the correct test time convention (fourth panel), approaching the reward of centralized training (black dashed line), while supervised training does not. The dashed line indicates the average reward of separate self-play training, and error bars represent $95 \%$ confidence intervals derived from independent replicates.} 
\label{traffic_fig}
\end{center}
\end{figure*}

We wish to use the insights from initialization in environments where function approximation (e.g. with deep neural networks) is required. However, if the policy is computed via function approximation, it is not clear how to `initialize' its value at particular states. Specifically, the policy at the small number of states in $\mathcal{D}$ can only be expected to generalize if the approximation captures the regularities of the game, which will only be true after some interaction in the environment. Therefore, we consider a procedure where consistency with $\mathcal{D}$ is enforced during training and smoothly decays over time.

We consider training with stochastic gradient descent using a loss function which is a linear combination of the likelihood of the data (a supervised objective) plus the policy gradient estimator of the reward in the game (we denote by $\mathcal{L}^{PG}$ the negative of this quantity). Formally, each agent receives rewards given by $$\mathcal{L}^{OSP}_i (\pi_i, \pi_{-i}) = \mathcal{L}^{PG}(\pi_i \mid \pi_{-i})  + \lambda  \sum_{(s,a) \in \mathcal{D}_i} \log(\pi_i(s, a))$$ with respect to the parameters $\theta_i$ of its policy.\footnote{Note this is different from reward shaping as the probably that a state is reached does not affect the supervised loss of the policy.}

We optimize the joint objective using MARL. At any time $t$ when making an update to the parameters of each agent $\theta_i$ we will take gradient steps of the form $$\nabla^{OSP}_t (\theta_i, \mathcal{D}) = \nabla_t^{PG} (\theta_i) + \lambda_t \nabla^{sup} (\theta_i, \mathcal{D}).$$ Where $\nabla^{PG}$ is our policy gradient and $\nabla^{sup}$ is the gradient of the supervised objective at $\theta_i.$\footnote{As with the best response dynamics above using a compound objective with a constant $\lambda$ can, in theory, introduce new equilibria during training. To be sure this does not occur we can anneal the weight on the supervised loss over time with $\lim_{t \to \infty} \lambda_t = 0.$ In practice, however, using a fixed $\lambda$ in our environments appeared to create policies that were still consistent with test time equilibria thus suggesting that if new equilibria were introduced they did not have large basins of attraction for our policy-gradient based learning procedures. }

Our main analysis is experimental and we use three environments from the literature: traffic, language games, and risky coordination. Our main results are:

\begin{claim}[Experiments 1,2,3]
OSP greatly increases the probability our agent learns a convention compatible with test time agents in situations where standard self-play by itself does not guarantee good test time performance and $\mathcal{D}$ is insufficient to learn a good policy by behavioral cloning alone.
\end{claim}

\begin{claim}[Experiment 3]
OSP can find conventions that have a small basin of attraction for MARL alone. Thus OSP can be used in situations where self-play will rarely find a test-time compatible convention.
\end{claim}

For all experiments, we represent the states using simple neural networks. The first two experiments have relatively low dimensional state representations so we use two layer MLPs with $128$ hidden units per layer and ELU non-linearities. Our third experiment has a grid structure so we represent the state using the convolutional neural network architecture from \cite{peysakhovich2017prosocial}. 

For RL training we use A3C using a multi-agent variant of the pytorch-a3c package \cite{pytorchaaac} run on $40$ threads. We use $20$ step returns and $\gamma = .99$. We use the Adam method for optimization \cite{kingma2014adam}. For OSP, we add the supervised term to the A3C loss with $\lambda=1$, using minibatches from $\mathcal{D}$ of size 20. Environment-specific parameters are detailed in the subsections below. In each experiment we compare the performance of OSP for various sizes of $\mathcal{D}$. We populate $\mathcal{D}$ with actions for all agents for states sampled at uniform intervals from true test time play.

\subsection{Traffic}
We first consider a multi-agent traffic navigation game inspired by \cite{sukhbaatar2016learning}. Each of $10$ agents begins at a random edge of a grid and can move around in each of the $4$ cardinal directions or stand still (see Figure \ref{traffic_fig}). Each agent is given a goal to reach. When the agent reaches the goal they receive a reward of $+1$ and then a new goal. If agents collide with another agent or wall they receive a negative reward ($-5$ for colliding with another agent, $-.1$ for colliding with a wall). Agents do not have full observation of the environment. Rather, they have access to the position of their goal relative to themselves and a local view of any other agents nearby. We train agents for a total of $800,000$ episodes.\footnote{We found it necessary to ramp the collision penalty linearly over the first $400,000$ episodes to avoid agents becoming stuck in the local minima of never moving.}

We train $20$ replicates and see that two incompatible conventions emerge. This can be seen in Figure \ref{traffic_fig} where we plot payoffs to an agent from one replicate paired with $9$ agents from another. We visualize the conventions by taking the empirical average action taken by any agent in any of the possible traffic coordinate locations (Figure \ref{traffic_fig} panel 3). We find that the two conventions that emerge are similar to the real world: either agents learn to drive on the left of the road or they learn to drive on the right.

We now train agents using OSP and insert them into environments with $9$ pre-converged agents. The test time payoffs to the transplanted agent for various sizes of $\mathcal{D}$ are shown in Figure \ref{traffic_fig} panel 4 top. The dashed line corresponds to the expected payoff of an agent trained using standard self-play (no observations). We see 20 observations ($2$ observations for each of the $10$) agents is sufficient to guide OSP to compatible conventions. The bottom panel shows that this is not enough data to train a good strategy via behavioral cloning alone (i.e. using just the supervised objective).

\subsection{Language}

\begin{figure}[ht!]
\begin{center}
\includegraphics[scale=.4]{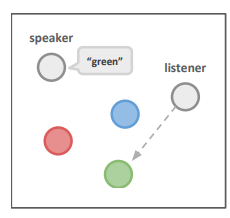}
\includegraphics[scale=.5]{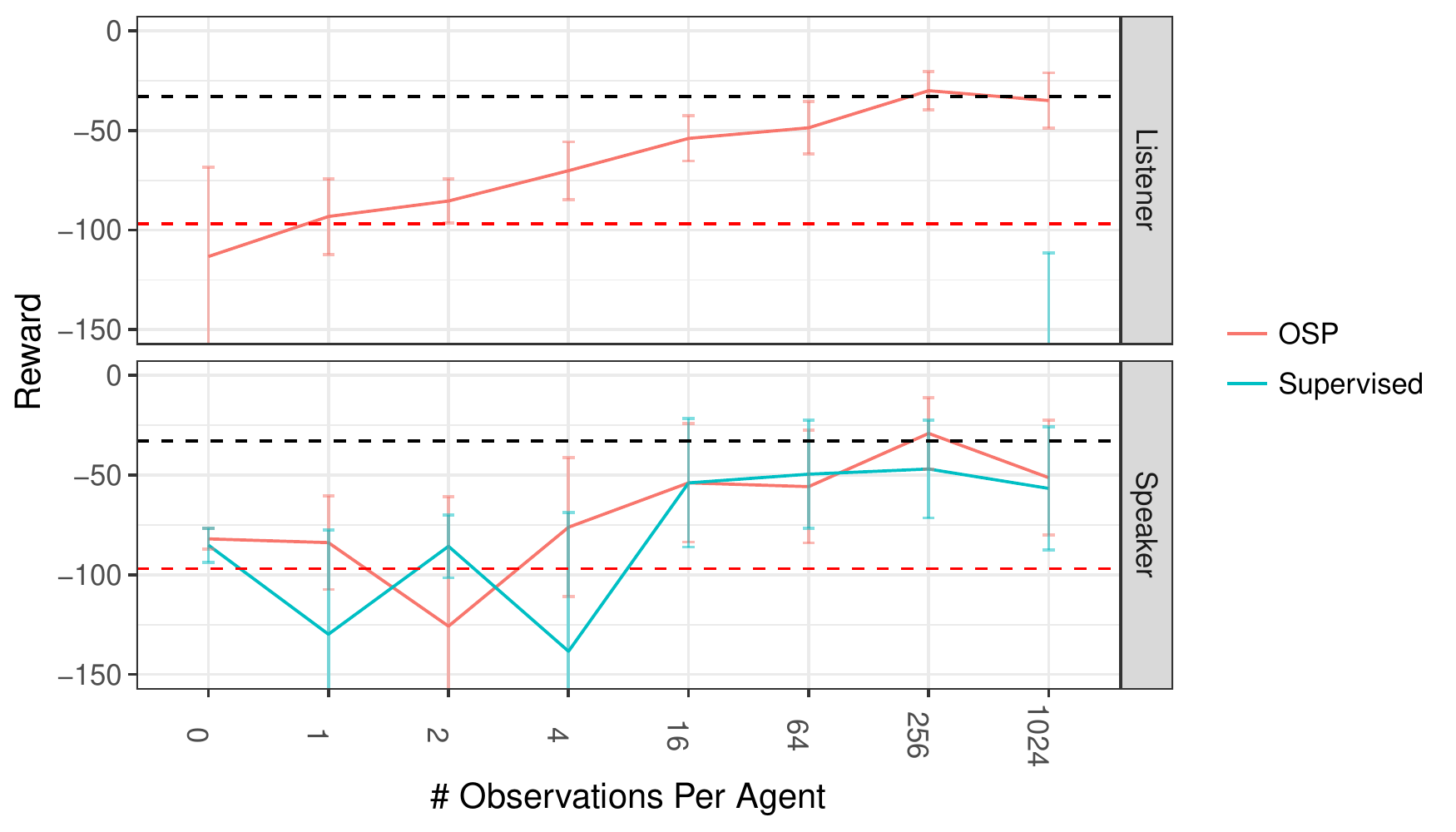}
\caption{In the speaker/listener particle environment (first panel taken from the original paper) when a pair are trained together they can reach high payoffs (black dashed line) but pairs trained separately with self-play (red dashed line) perform quite poorly. OSP leads to agents that reach high payoffs with their test time partner. Supervision alone is insufficient to construct good Listener agents (plot is below the y axis for N < 1024). Error bars represent confidence intervals derived from independent replicates.} 
\label{speaker_listener_fig}
\end{center}
\end{figure}

An important type of convention is language. There is a resurgence of interest in the deep RL community in using communication games to construct agents which are able to communicate \cite{jorge2016learning,foerster2016learning,lazaridou2017multi,lowe2017multi}. 

We now apply OSP to the cooperative communication task in the particle environment studied by \cite{lowe2017multi} and \cite{mordatch2017emergence}. In this environment there are two agents, a speaker and a listener, and three landmarks (blue, red, or green). One of the landmarks is randomly chosen as a goal and the reward for both agents at each step is equal to the distance of the listener from the landmark. However, which landmark is the goal during a particular episode is only known to the speaker who is able to produce a communication output from a set of $20$ symbols. To solve the cooperation task, agents thus need to evolve a simple `language'. This language only requires $3$ symbols to be used, but this still allows for at least $6840$ incompatible conventions (one symbol per landmark).

In this experiment we use a lower discount factor of $\gamma=.8$ and as suggested by \cite{lowe2017multi} we also use a centralized critic. It was shown in prior work that if artificial agents learn language by self-play they can learn arbitrary languages which may not be compatible with new partners \cite{lazaridou2017multi}. Indeed, when we pair two agents who were trained separately they clearly do not speak the same language - i.e. they cannot communicate and so achieve low payoffs.

We look at the effect of adding observational data to the training of either a speaker or listener (we train a total of $135$ replicates to convergence). In the case of the speaker (whose policy is a simple map from goals to communication symbols) supervision is sufficient to learn the a good test-time language. However, pure behavioral cloning fails catastrophically for the listener. Again, OSP with a relatively small number of observations is able to achieve high payoffs (Figure \ref{speaker_listener_fig}).

\subsection{Risky Coordination}
\begin{figure}[ht!]
\begin{center}
\includegraphics[scale=.3]{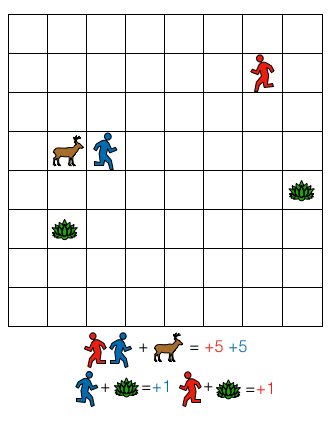}  \\
\includegraphics[scale=.4]{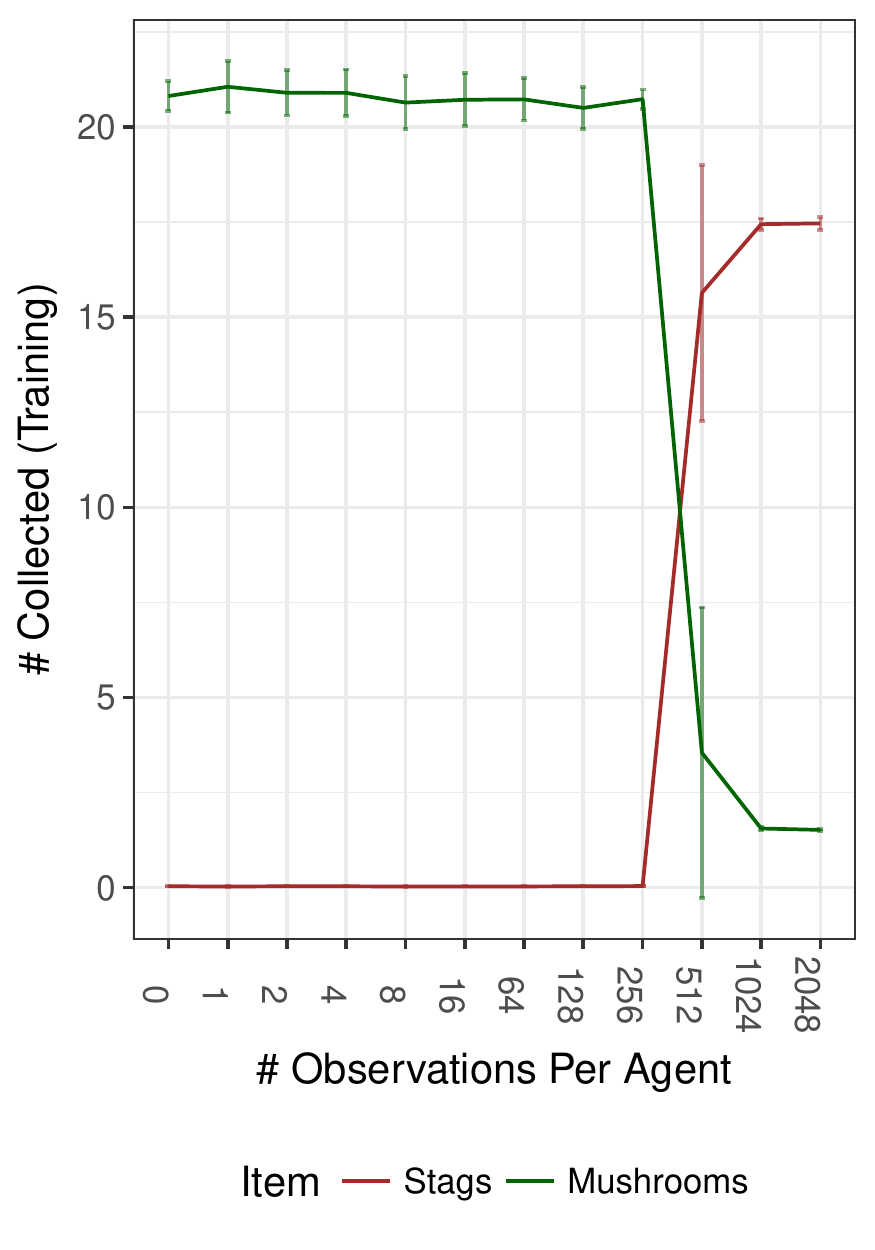}
\includegraphics[scale=.4]{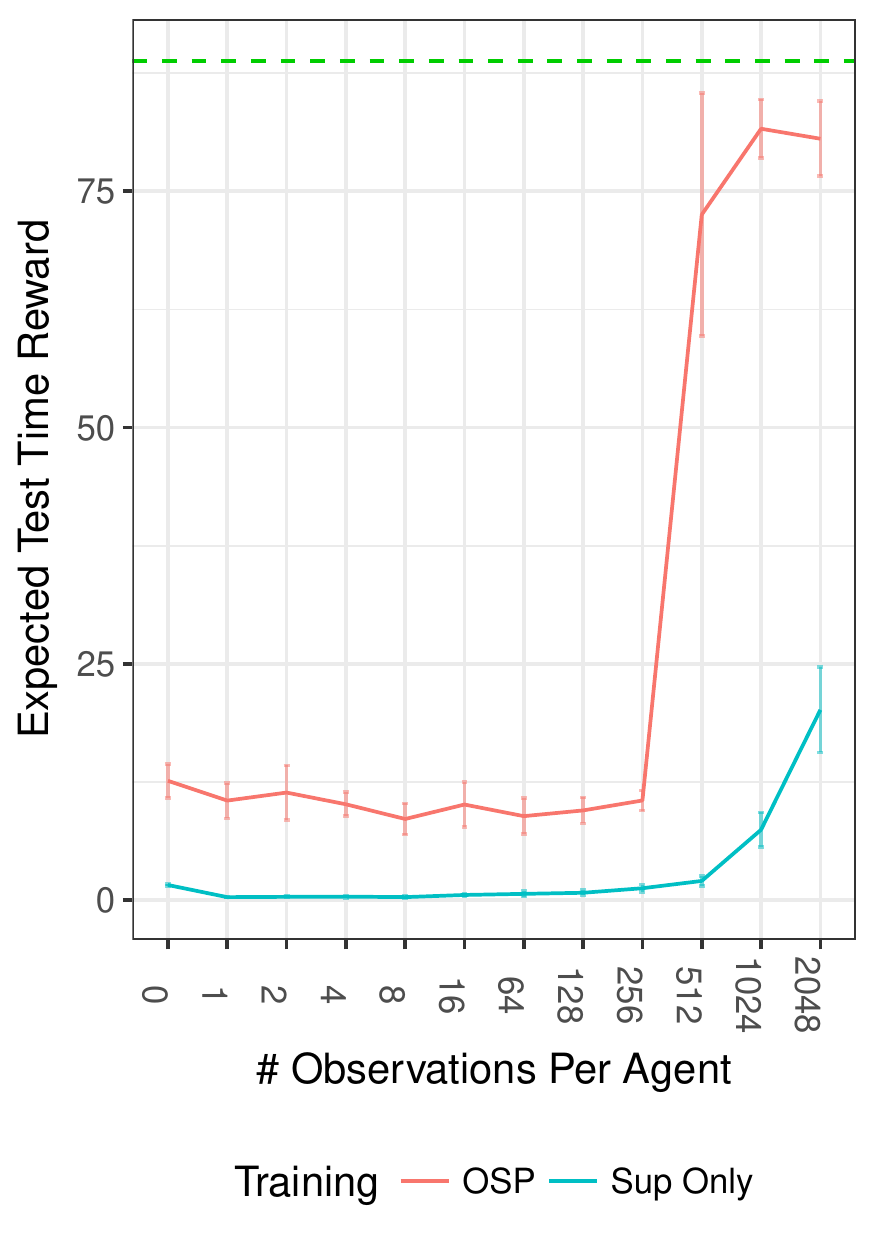}

\caption{In the Stag Hunt independent MARL typically converges to the safe equilibrium. Thus, an agent trained via MARL cannot coordinate when matched with a test-time partner who choose to hunt. However OSP, with samples from the Hunt equilibrium, can converge to the Hunting convention during training reaching a payoff almost as high as two Hunters trained together (green line). Error bars represent confidence intervals derived from independent replicates.}
\label{stag_fig}
\end{center}
\end{figure}
We now consider a risky coordination game known as the Stag Hunt. The matrix game version of the Stag Hunt has both agents choosing either to Hunt (an action that require coordination) or Forage (a safe action). Foraging yields a sure (low) payoff whereas Hunting yields a high payoff if the other agent chooses to Hunt also and a low payoff if one shows up alone. It is known that in both matrix and Markov versions of Stag Hunt games many standard self-play based algorithms yield agents that converge to the inefficient equilibrium in which both agents choose safe actions. This happens because while our partner is not hunting effectively (i.e. early in training), the payoff to hunting ourselves is quite low. Thus, the basin of attraction of joint hunting is much smaller than the basin of attraction of both foraging. 

This situation is different from the ones in traffic and language: here there are multiple conventions (hunt or forage) but they are not payoff equivalent (hunting is better) nor do they have similar sized basins of attraction (hunting is very difficult to find via standard independent MARL). 

We use the Markov version of Stag Hunt introduced by \cite{peysakhovich2017prosocial} where two agents live on a $8 \times 8$ grid. Two plants and a stag are placed at random locations. If an agent moves over a plant, it receives a reward of $1$. Stepping on the stag gives a reward of $5$ to both players if they step on it simultaneously, otherwise there is no reward. When either a plant or stag is stepped on, it restarts in a new location. Games last $100$ rounds. 

We start by constructing a test time hunting partner by inducing joint hunting strategies in $40$ replicates. Because MARL by itself does not find hunt equilibria, we construct a hunting partner by training an agent under a modified payoff structure (payoff of $0$ for plants; payoff of $0.1$ for unilateral hunting). 

We then test whether we can train agents in the \textit{original} game who can coordinate with test time partners that hunt. We use OSP with varying amounts of data from the hunting agents. We see that with moderate amounts of data OSP often converges to the hunting convention at test time even though two agents trained together using independent MARL fail to find the high payoff joint stag equilibrium in \textit{any} of the replicates. As a result, OSP outperforms even centralized self-play because the observations of the risky partner guide the agent to a better equilibrium. As with the traffic and language environments above we see that pure behavioral cloning is insufficient to construct good test time strategies (Figure \ref{stag_fig}).

\section{Conclusion}
Conventions are an important part of social behavior and many multi-agent environments support multiple conventions as equilibria. If we want to construct artificial agents that can adapt to people (rather than requiring people to adapt to them) these agents need to be able to act according to the existing social conventions. In this paper we have discussed how a simple procedure that combines small amounts of imitation learning with self-play can lead to agents that can learn social conventions.

There are many open questions remaining in the study of conventions and building agents that can learn them quickly. OSP uses a straightforward combination of RL and behavioral cloning. It would be interesting to explore whether ideas from the learning with expert demonstrations literature \cite{hester2017learning}. In addition, OSP follows current deep RL paradigms splits strategy construction into a training and a test phase. An interesting extension is to consider the OSP training strategies can be fine-tuned during test time.

We have focused on situations where agents have no incentive to deviate from cooperation and only need to learn correct conventions. An important future direction is considering problems where agents have partially misaligned incentives but, in addition to just solving the social dilemma, must also coordinate on a convention \cite{kleiman2016coordinate,leibo2017multi,lerer2017maintaining,foerster2017learning,peysakhovich2017consequentialist}.

There is a large recent interest in hybrid systems which include both human and artificially intelligent participants \cite{shirado2017locally}. Thus, another key extension of our work is to understand whether techniques like OSP can construct agents that can interact with humans in more complex environments.

\bibliography{20190304_conventions_AIESFinal.bbl}
\bibliographystyle{ACM-Reference-Format}

\newpage

\section{Appendix}
\subsection{Relationship Between Markov Strategic Complements and Strategic Complements}
The original definition of strategic complements comes from games with continuous actions used to model multiple firms in a market. In the simplest example we have multiple firms which produce units of goods $(x_1, x_2, \dots, x_n)$. The revenue function of each firm $i$ is $R_i (x_i, x_{-i})$ where $R_i$ is smooth, strictly concave, increasing and has $R_i (x_i = 0, x_{-i}) = 0$. The goods are strategic complements if $\dfrac{\partial^2 R_i}{\partial x_i \partial x_j} > 0$, in other words goods are strategic complements if ``more `aggressive' play... by one firm... raises the marginal profitabilities [of the others].''  \cite{bulow1985multimarket} Firms have costs of production given by $c_i (x_i)$ which has $c(0) = 0$, $c'(0)= 0$, is convex, and increasing. Thus each firm's objective function is $$R_i (x_i, x_{-i}) - c_i (x_i)$$

If firm $-i$ is producing $x_{-i}$ then firm $i$'s best response $x^*_i (x_{-i})$ sets $$\dfrac{\partial R_i (x^*_i (x_{-i}), x_{-i})}{\partial x_i} = \dfrac{\partial c_i (x^*_i (x_{-i})}{\partial x_i}.$$ Given the definition of strategic complements above this means that $\dfrac{\partial x^*_i}{\partial x_{j}} > 0$ for all other firms $j$. 

Strategic complements implies our Markov strategic complements in a matrix game with multiple equilibria (since any firm changing their production level higher or lower causes other firms to also want to change their production). Markov strategic complements is weaker than strategic complements in matrix games since it only pins down how best responses to shift when others change to equilibrium actions rather than any action shift (though if action spaces in each state were totally ordered one could amend the definition to keep all of the properties). 

\subsection{Proof of Main Theorem}

\ \\
 \textbf{Lemma 1:} In a Markov strategic complements (MSC) game, any policy $\pi$ in the basin of attraction of an equilibrium $A$ remains there under observational initialization, i.e. $\pi_i \in \Pi^A \implies \bar{\pi}_i(\mathcal{D}_A) \in \Pi^A$. 
\\

We define the operator $BR^{(k)}$ as $k$ iterations of the best response operator, $$BR^{(k)}(\pi_i) = BR^i (BR^{-i} ( BR^{i} (\dots BR^{-i} (\pi_i)))).$$

Consider an initial policy $\pi_i \in \Pi^A$ for some equilibrium A. There exists $k_c$ such that $BR^{(k_c)} = A_i$. Now consider an observationally initialized policy $\bar{\pi}_i(\mathcal{D}_A)$ for some dataset $\mathcal{D}$ drawn from $A_i$. By definition, this implies that $\bar{\pi}_i \succsim_{A} \pi_i$. Now, since the game is MSC, $$BR^{-i} (\bar{\pi}_i(\mathcal{D}_A)) \succsim_{A} BR^{-i} (\pi_i).$$ By repeated application of the MSC property, we find that for all $k$, $$BR^{(k)} (\bar{\pi}_i(\mathcal{D}_A)) \succsim_{A} BR^{(k)} (\pi_i).$$ To conclude, we note that $BR^{(k_c)}(\bar{\pi}_i^0(\mathcal{D}_A)) \succsim_{A} A_i$, which implies $BR^{(k_c)}(\bar{\pi}_i(\mathcal{D}_A)) = A_i$.
\\

\textbf{Lemma 2:} In a MSC game with a finite number of states, there exists a state $s$ such that for any $\mathcal{D}_A$ that contains the state-action pair $(s, A_i(s))$, there is a policy not in the basin of attraction of $A$ but which enters the basin of attraction of $A$ under observational initialization.
\\

Consider a policy $\pi_i^{0} \notin \Pi_A$, and order the states lexicographically $(s_1, s_2, \dots, s_M)$. Now consider the sequence of policies $\pi_i^{k}$ where $\pi_i^{k} (s_l) = A_i (s_l)$ for $l < k$ and $\pi_i^{k} (s_l) = \pi^{0} (s_l)$ for $k \geq l.$ We know that $\pi_i^M = A_i \in \Pi_A$, therefore there exists some $t$ such that $\pi_i^t \notin \Pi_A$ and $\pi_i^{t+1} \in \Pi_A$. Now, consider a dataset $\mathcal{D}_A$ containing the state-action pair $(s_{t+1}, A_i(s_{t+1}))$. Then $\bar{\pi_i^t}(\mathcal{D}_A) \succsim_{A} \pi^{t+1}$. As discussed in the last section, if $\pi^{t+1} \in \Pi_A$ and $\bar{\pi_i^t}(\mathcal{D}_A) \succsim_{A} \pi^{t+1}$, then $\bar{\pi_i^t}(\mathcal{D}_A) \in \Pi_A$. Therefore, for any dataset containing $s_{t+1}$, the policy $\pi_i^t$ enters the basin of attraction of $A$ under observational initialization.

\end{document}